\documentclass[10pt,twocolumn,letterpaper]{article}

\usepackage{iccv}
\usepackage{times}
\usepackage{epsfig}
\usepackage{graphicx}
\usepackage{amsmath}
\usepackage{amssymb}
\usepackage{multirow}

\usepackage[breaklinks=true,bookmarks=false]{hyperref}

\iccvfinalcopy 

\def\iccvPaperID{****} 

\begin{document}

\title{Be Your Own Teacher: Improve the Performance of Convolutional Neural Networks via Self Distillation}

\author{Linfeng Zhang\\
Tsinghua University\\
{\tt\small zhanglinfeng1997@outlook.com}
\and
Jiebo Song\\
IIISCT\\
{\tt\small songjb@iiisct.com}
\and
Anni Gao\\
IIISCT\\
{\tt\small gaoan@iiisct.com}
\and
Jingwei Chen\\
Hisilicon\\
{\tt\small jean.chenjingwei@hisilicom}
\and
Chenglong Bao\\
Tsinghua University\\
{\tt\small clbao@mail.tsinghua.edu.cn}
\and
Kaisheng Ma\\
Tsinghua University\\
{\tt\small kaisheng@mail.tsinghua.edu.cn}}

\maketitle

\begin{abstract}

Convolutional neural networks have been widely deployed in various application scenarios.
In order to extend the applications' boundaries to some accuracy-crucial domains, researchers have been investigating approaches to boost accuracy through either deeper or wider network structures, which brings with them the exponential increment of the computational and storage cost, delaying the responding time.

In this paper, we propose a general training framework named self distillation, which notably enhances the performance (accuracy) of convolutional neural networks through shrinking the size of the network rather than aggrandizing it.
Different from traditional knowledge distillation - a knowledge transformation methodology among networks, which forces student neural networks to approximate the softmax layer outputs of pre-trained teacher neural networks, the proposed self distillation framework distills knowledge within network itself. The networks are firstly divided into several sections. Then the knowledge in the deeper portion of the networks is squeezed into the shallow ones.
Experiments further prove the generalization of the proposed self distillation framework:  enhancement of accuracy at average level is 2.65\%, varying from 0.61\% in ResNeXt as minimum to 4.07\% in VGG19 as maximum.
In addition, it can also provide flexibility of depth-wise scalable inference on resource-limited edge devices.
Our codes will be released on github soon.
\end{abstract}

 \begin{figure}
  \centering
  \includegraphics[width=6.94cm,height=11.6cm]{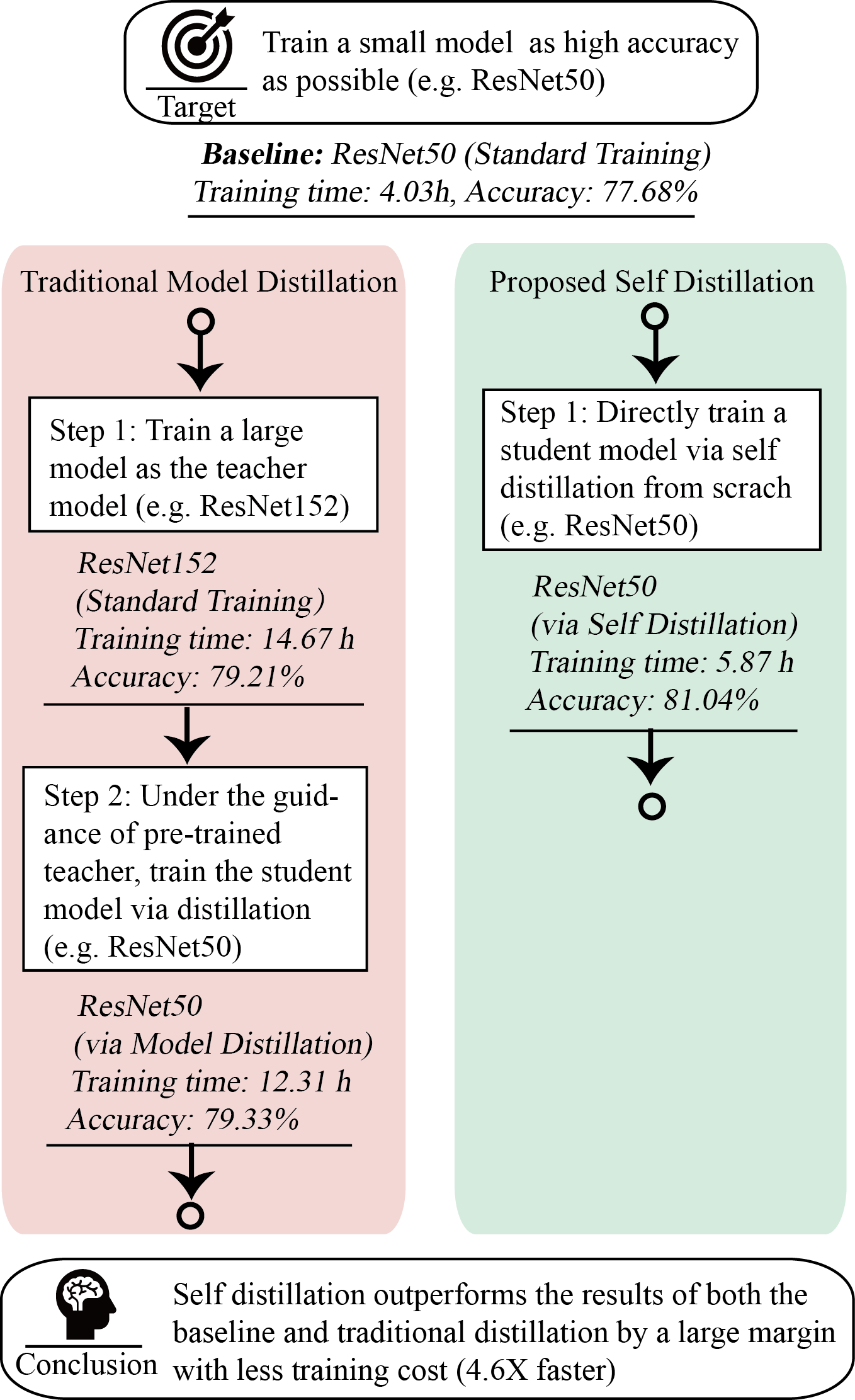}
  \caption{Comparison of training complexity, training time, and accuracy between traditional distillation and proposed self distillation (reported on CIFAR100).}
  \label{Compare}
\end{figure}

\section{Introduction}

With the help of convolutional neural networks, applications such as image classification \cite{krizhevsky2012imagenet,simonyan2014very} ,object detection \cite{liu2016ssd}, and semantic segmentation \cite{dou20163d, yu2017volumetric} are developing at an unprecedented speed nowadays.
Yet, in some applications demanding intolerate errors, such as automated driving and medical image analysis, prediction and analysis accuracy needs to be further improved, while at the same time, shorter response time is required. This leads to tremendous challenges on current convolutional neural networks.
Traditional methods were focused on either performance improvement or reduction of computational resources (thus response time).
On the one hand, for instance, ResNet 150 or even larger ResNet 1000 have been proposed to improve very limited performance margin but with massive computational penalty.
On the other hand, with a pre-defined performance lost compared with best effort networks, various techniques have been proposed to reduce the computation and storage amount to match the limitations brought by hardware implementation. Such techniques include lightweight networks design \cite{iandola2016squeezenet,howard2017mobilenets}, pruning \cite{han2015deep,han2015learning} and quantization \cite{courbariaux2015binaryconnect,rastegari2016xnor}.
Knowledge Distillation (KD) \cite{hinton2015distilling} was one of the available approaches, or even regarded as a trick, to achieve model compression.


As one of the popular compression approaches, knowledge distillation \cite{hinton2015distilling} is inspired by knowledge transfer from teachers to students. Its key strategy is to orientate compact student models to approximate over-parameterized teacher models. As a result, student models can gain significant performance boost which is sometimes even better than that of teacher's. By replacing the over-parameterized teacher model with a compact student model, high compression and rapid acceleration can be achieved.
However, glories come with remaining problems. The first setback is low efficiency on knowledge transfer, which means student models scarcely exploit all knowledge from teacher models. A distinguished student model which outperforms its teacher model remains rare.
Another barrier is how to design and train proper teacher models. The existing distillation frameworks require substantial efforts and experiments to find the best architecture of teacher models, which takes a relatively long time.

 \begin{figure*}
  \centering
  \includegraphics[width=16.29cm,height=5.97cm]{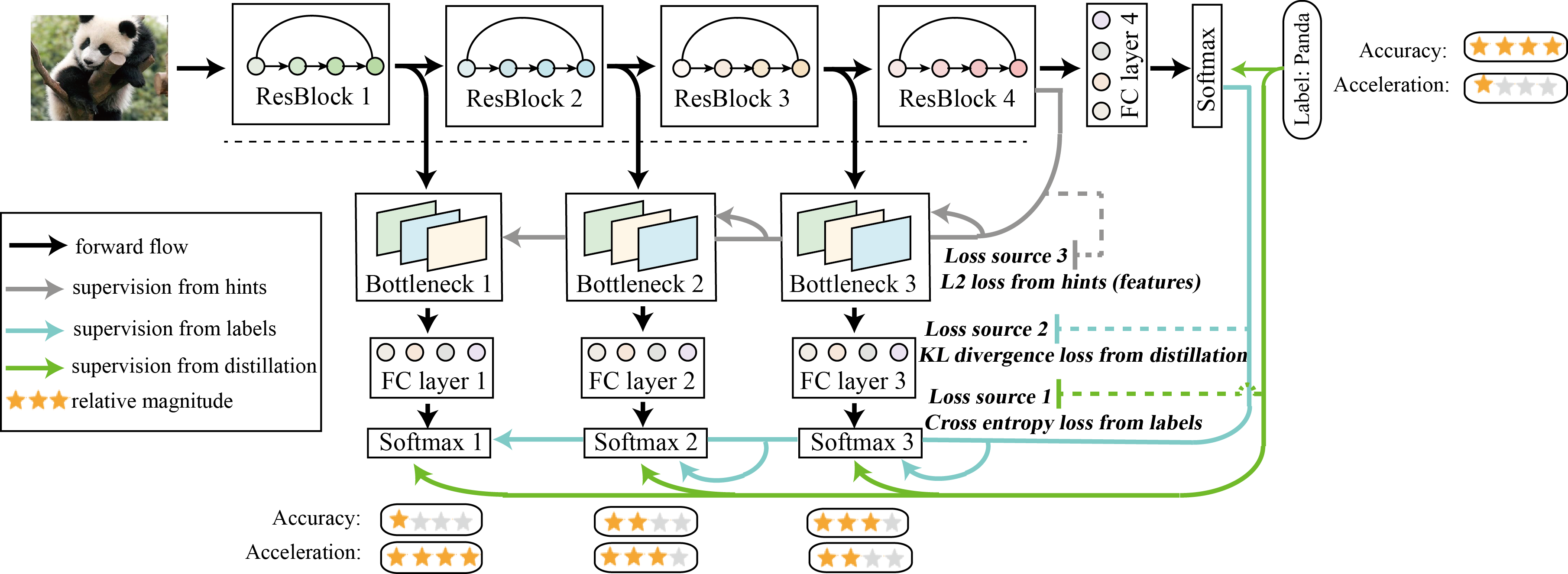}
  \caption{This figure shows the details of a ResNet equipped with proposed self distillation. (i) A ResNet has been divided into four sections according to their depth. (ii) Additional bottleneck and fully connected layers are set after each section, which constitutes multiple classifiers. (iii) All of the classifiers can be utilized independently, with different accuracy and response time. (iv) Each classifier is trained under three kinds of supervision as depicted. (v) Parts under the dash line can be removed in inference.}
  \label{BigStructure}
\end{figure*}

As shown in Figure \ref{Compare}, in order to train a compact model to achieve as high accuracy as possible and to overcome the drawbacks of traditional distillation, we propose a novel self distillation framework.
Instead of implementing two steps in traditional distillation, that is first, to train a large teacher model, and second, to distill the knowledge from it to the student model, we propose a one-step self distillation framework whose training points directly at the student model.
The proposed self distillation not only requires less training time (from 26.98 hours to 5.87 hours on CIFAR100, a 4.6X time training shorten time), but also can accomplish much higher accuracy (from 79.33\% in traditional distilllaitn to 81.04\% on ResNet50).



In summary, we make the following principle contributions in this paper:
\begin{itemize}
  \item Self distillation improves the performance of convolutional neural networks by a large margin at no expense of response time. 2.65\% accuracy boost is obtained on average, varying from 0.61\% in ResNeXt as minimum to 4.07\% in VGG19 as maximum.

  \item Self distillation provides a single neural network executable at different depth, permitting adaptive accuracy-efficiency trade-offs on resource-limited edge devices.

  \item  Experiments for five kinds of convolutional neural networks on two kinds of datasets are conducted to prove the generalization of this technique.
\end{itemize}

The rest of this paper is organized as follows. Section ~\ref{related work} introduces the related work of self distillation. Section ~\ref{self distillation} demonstrates the formulation and detail of self distillation. Section ~\ref{experiments} shows the experiments results on five kinds of convolutional networks and two kinds of datasets. Section \ref{discussion} explains the reason why self distillation works. Finally, a conclusion is brought forth in section ~\ref{conclusion}.

\section{Related Work\label{related work}}
\textbf{Knowledge distillation:}
knowledge distillation is one of the most popular techniques used in model compression \cite{buciluǎ2006model,hinton2015distilling}.
 A large quantity of approaches have been proposed to reinforce the efficiency of student models' learning capability. Romero \emph{et al.} firstly put forward FitNet in which the concept of hint learning was proposed, aiming at reducing the distance between feature maps of students and teachers \cite{romero2014fitnets}.
 Agoruyko \emph{et al.} \cite{zagoruyko2016paying} considered this issue from the perspective of attention mechanism, attempting to align the features of attention regions.
 Furthermore, some researchers extended knowledge distillation to generative adversarial problem \cite{shen2018meal,liu2018ktan}.

In the other domains, knowledge distillation also shows its potential.
Furlanello \emph{et al.} interactively absorbed the distillated student models into the teacher model group, through which the better generalization ability on test data is obtained \cite{furlanello2018born}.
Bagherinezhad \emph{et al.} applied knowledge distillation to data argumentation, increasing the numerical value of labels to a higher entropy \cite{bagherinezhad2018label}. Papernot \emph{et al.} regarded knowledge distillation as a tool to defend adversarial attack \cite{papernot2016distillation}, and Gupta \emph{et al.}, using the same methods, transferred the knowledge among data in different modals \cite{gupta2016cross}.

As shown above, in general, teacher models and student models work in their own ways respectively, and knowledge transfer flows among different models.
In contrast, student and teacher models in our proposed self distillation method come from the same convolutional neural networks.

\textbf{Adaptive Computation:}
Some researchers incline to selectively skip several computation procedures to remove redundancy.
Their work can be witnessed from three different angles: layers, channels and images.

\emph{Skipping some layers in neural networks}. Huang \emph{et al.} proposed random layer-wise dropout in training \cite{huang2016deep}.
Some researchers extended this idea to inference.
Wang \emph{et al.} and Wu \emph{et al.} further extended the layer-wise dropout from training to inference by introducing additional controller modules or gating functions based on the current input \cite{wu2018blockdrop,wang2018skipnet}.
Another extension of the layer-wise dropout solutions is to design early-exiting prediction branches to reduce the average execution depth in inference \cite{huang2017multi,amthor2016impatient,veit2017convolutional,kuen2018stochastic}.


\emph{Skipping some channels in neural networks}. Yu \emph{et al.} proposed switchable batch normalization to dynamically adjust the channels in inference \cite{yu2018slimmable}.

\emph{Skipping less important pixels of the current input images}.
Inspired by the intuition that neural networks should focus on critical details of input data \cite{bahdanau2014neural}, reinforcement learning and deep learning algorithms are utilized to identify the importance of pixels in the input images before they are feed into convolutional neural networks \cite{mnih2014recurrent,fu2017look}.

\textbf{Deep Supervision:}
Deep supervision is based on the observation that classifiers trained on highly discriminating features can increase the performance in inference \cite{lee2015deeply}.
In order to address the vanishing gradient problem, additional supervision is added to train the hidden layers directly. For instance, significant performance gain has been observed in tasks like image classification \cite{lee2015deeply}, objection detection \cite{lin2018focal,lin2017feature,liu2016ssd}, and medical images segmentation \cite{yu2017volumetric,dou20163d}.

The multi-classifier architecture adopted in the proposed self distillation framework is similar to deeply supervised net \cite{lee2015deeply}.
The main difference in self distillation is that shallow classifiers are trained via distillation instead of only labels, which leads to an obvious higher accuracy supported by experiments results.

\begin{table*}[htbp]
\begin{center}
\begin{tabular}{|c|c|c|c|c|c|c|c|c|}
\hline
Neural Networks& Baseline & Classifier 1/4 &Classifier 2/4 &Classifier3/4 &Classifier 4/4 & Ensemble\\
\hline
VGG19(BN)&64.47 &\textcolor{red}{63.59} &67.04  &68.03 &67.73& 68.54 \\
\hline
ResNet18&77.09 &\textcolor{red}{67.85} &\textcolor{red}{74.57}  &78.23 &78.64  &79.67 \\
\hline
ResNet50&77.68 &\textcolor{red}{68.23}  &\textcolor{red}{74.21}  &\textcolor{red}{75.23} &80.56  &81.04 \\
\hline
ResNet101&77.98  &\textcolor{red}{69.45}  &\textcolor{red}{77.29} &81.17  &81.23&82.03 \\
\hline
ResNet152&79.21  &\textcolor{red}{68.84}  & \textcolor{red}{78.72} &81.43 & 81.61 &82.29 \\
\hline
ResNeXt29-8& 81.29&\textcolor{red}{71.15}  &\textcolor{red}{79.00} &81.48 &81.51 &81.90 \\
\hline
WideResNet20-8& 79.76 &\textcolor{red}{68.85}  & \textcolor{red}{78.15} &80.98 &80.92  &81.38 \\
\hline
WideResNet44-8& 79.93 &\textcolor{red}{72.54}  &81.15  &81.96 &82.09  &82.61 \\
\hline
WideResNet28-12& 80.07 &\textcolor{red}{71.21}  &80.86 &81.58& 81.59 & 82.09\\
\hline
PyramidNet101-240& 81.12&\textcolor{red}{69.23} & \textcolor{red}{78.15} &80.98   &82.30  &83.51\\
\hline
\end{tabular}
\end{center}
\caption{Experiments results of accuracy (\%) on CIFAR100 (the number marked in red is lower than its baseline).}
\label{table1}
\end{table*}

\begin{table*}[htbp]
\begin{center}
\begin{tabular}{|c|c|c|c|c|c|c|c|c|}
\hline
Neural Networks& Baseline & Classifier 1/4 &Classifier 2/4 &Classifier 3/4 &Classifier 4/4 & Ensemble\\
\hline
VGG19(BN)&70.35&\textcolor{red}{42.53}&\textcolor{red}{55.85}&71.07&72.45& 73.03\\
\hline
ResNet18&68.12 &\textcolor{red}{41.26} &\textcolor{red}{51.94} &\textcolor{red}{62.29} &69.84 &68.93 \\
\hline
ResNet50&73.56&\textcolor{red}{43.95} &\textcolor{red}{58.47} &\textcolor{red}{72.84} &75.24 &74.73\\
\hline
\end{tabular}
\end{center}
\caption{Experiments results of top-1 accuracy (\%) on ImageNet (the number marked in red is lower than its baseline).}
\label{imagenet}
\end{table*}

\section{Self Distillation\label{self distillation}}

In this section, we put forward self distillation techniques as depicted in Figure \ref{BigStructure}.
We construct the self distillation framework in the following ways of thinking:
To begin with, the target convolutional neural network is divided into several shallow sections according to its depth and original structure. For example, ResNet50 is divided into 4 sections according to ResBlocks.
Secondly, a classifier, combined with a bottleneck \cite{he2016deep} layer and a fully connected layer which are only utilized in training and can be removed in inference, is set after each shallow section. The main consideration of adding the bottleneck layer is to mitigate the impacts between each shallow classifier, and to add L2 loss from hints.
While in training period, all the shallow sections with corresponding classifiers are trained as student models via distillation from the deepest section, which can be conceptually regarded as the teacher model.

In order to improve the performance of the student models, three kinds of losses are introduced during training processes:
\begin{itemize}
  \item Loss Source 1: Cross entropy loss from labels to not only the deepest classifier, but also all the shallow classifiers.
  It is computed with the the labels from the training dataset and the outputs of each classifer's softmax layer.
  In this way, the knowledge hidden in the dataset are introduced directly from labels to all the classifiers.

  \item Loss Source 2: KL (Kullback-Leibler) divergence loss under teacher's guidance.
  The KL divergence is computed using softmax outputs between students and teachers, and introduced to the softmax layer of each shallow classifier.
  By introducing KL divergence, the self distillation framework affects the teacher's networks, the deepest one, to each shallow classifier.

  \item Loss Source 3: L2 loss from hints.
  It can be obtained through computation of the L2 loss between features maps of the deepest classifier and each shallow classifier. By means of L2 loss, the inexplicit knowledge in feature maps is introduced to each shallow classifier's bottleneck layer, which induces all the classifiers' feature maps in their bottleneck layers to fit the feature maps of the deepest classifier.

\end{itemize}

For that all the newly added layers (parts under the dash line in Figure \ref{BigStructure}) are only applied during training. They exert no influence during inference.
Adding these parts during inference provides another option for dynamic inference for energy constrained edge devices.


\begin{table*}[h]
\begin{center}
\begin{tabular}{|c|c|c|c|c|c|c|c|}
\hline
Teacher Model& Student Model & Baseline &KD \cite{hinton2015distilling} & FitNet \cite{romero2014fitnets} & AT \cite{zagoruyko2016paying} & DML \cite{zhang2018deep} & Our approach \\
\hline
ResNet152& ResNet18 & 77.09 & 77.79 & 78.21  & 78.54 & 77.54 & 78.64 \\
\hline
ResNet152& ResNet50 & 77.68 & 79.33 & 80.13 &79.35 &  78.31&   80.56\\
\hline
WideResNet44-8& WideResNet20-8 &79.76  &79.80  & 80.48 &80.65  & 79.91 & 80.92 \\
\hline
WideResNet44-8& WideResNet28-12 &80.07  & 80.95& 80.53 &81.46  &80.43  &  81.58\\
\hline
\end{tabular}
\end{center}
\caption{ Accuracy (\%) comparison with traditional distillation on CIFAR100.}
\label{table2}
\end{table*}

\begin{table*}[h]
\begin{center}
\begin{tabular}{|c|c|c|c|c|c|c|}
\hline
Neural Networks& Method & Classifier 1/4 &Classifier 2/4 &Classifier3/4 &Classifier 4/4 & Ensemble\\
\cline{1-7}
\multirow{2}{*}{ResNet18}&DSN &67.23 & 73.80 &77.75 &78.38  &79.27 \\
\cline{2-7}
&Our approach & 67.85 &74.57  &78.23 &78.64  &79.67 \\
\cline{1-7}
\multirow{2}{*}{ResNet50}&DSN &67.87 &73.80&74.54&80.27&80.67\\
\cline{2-7}
&Our approach &68.23  & 74.21 & 75.23 & 80.56 &81.04 \\
\cline{1-7}
\multirow{2}{*}{ResNet101}&DSN&68.17&75.43&80.98&81.01&81.72 \\
\cline{2-7}
&Our approach &69.45&77.29&81.17&81.23&82.03\\
\cline{1-7}
\multirow{2}{*}{ResNet152}&DSN&67.60&77.04&81.06&81.35&81.83 \\
\cline{2-7}
&Our approach &68.84&78.72&81.43&81.61&82.29\\
\cline{1-7}
\end{tabular}
\end{center}
\caption{Accuracy (\%) comparison with deeply supervised net \cite{lee2015deeply} on CIFAR100.}
\label{table3}
\end{table*}

\subsection{Formulation}
 Given $N$ samples $X = \{x_i\}_{i=1}^{N}$ from $M$ classes, we denote the corresponding label set as $Y = \{y_i\}_{i=1}^{M}$, $y_i \in \{1,2,...,M\}$. Classifiers (the proposed self distillation has multiple classifiers within a whole network) in the neural network are denoted as $\Theta = \{\theta_{i/C}\}_{i=1}^{C}$, where $C$ denotes the number of classifiers in convolutional neural networks. A softmax layer is set after each classifier.

\begin{equation}
q_i^c=\frac{\exp{(z_i^c/T)}}{\Sigma_j^c\exp{(z_j^c/T)}}
\label{equation1}
\end{equation}

Here $z$ is the output after fully connected layers. $q_i^c \in \mathbb{R^M}$ is the $i_{th}$ class probability of classifier $\theta_{c/C}$. $T$, which is normally set to 1, indicates the temperature of distillation~\cite{hinton2015distilling}. A larger T makes the probability distribution softer.

\subsection{Training Methods}
In self distillation, the supervision of each classifier $\theta_{i/C}$ except for the deepest classifier comes from three sources. Two hyper-parameters $\alpha$ and $\lambda$ are used to balance them.

\begin{equation}
(1-\alpha) \cdot CrossEntropy(q^i,y)
\label{equation2}
\end{equation}


The first source is the cross entropy loss computed with $q^i$ and labels $Y$. Note that $q^i$ denotes the softmax layer's output of classifier $\theta_{i/C}$ .


\begin{equation}
\alpha \cdot KL(q^i,q^C)
\label{equation3}
\end{equation}

The second source is the Kullback-Leibler divergence between $q^i$ and $q^C$. We aim to make shallow classifiers approximate the deepest classifier, which indicates the supervision from distillation. Note that $q^C$ means the softmax layer's output of the deepest classifier.

\begin{equation}
\lambda \cdot \| F_{i} - F_{C} \|_{2}^2
\label{equation4}
\end{equation}

The last supervision is from the hint of the deepest classifier. A hint is defined as the output of  teacher models hidden layers, whose aim is to guide the student models' learning \cite{romero2014fitnets}. It works by decreasing the distance between feature maps in shallow classifiers and in the deepest classifier. However, because the feature maps in different depth have different sizes, extra layers should be added to align them. Instead of using a convolutional layer \cite{romero2014fitnets}, we use a bottleneck architecture which shows positive effects on model's performance. Note that $F_{i}$ and $F_{C}$ denote features in the classifier $\theta_{i}$ and features in the deepest classifier $\theta_{C}$ respectively.

To sum up, the loss function of the whole neural networks consists of the loss function of each classifier, which can be written as:

\begin{equation}
\begin{aligned}
loss &= \sum_i^C loss_i
\\&= \sum_i^C\Big((1-\alpha)\cdot CrossEntropy (q^i,y)
\\&+ \alpha \cdot KL(q^i,q^C) + \lambda \cdot ||F_i - F_C||^2_2 \Big)
\end{aligned}
\label{equation5}
\end{equation}

Note that $\lambda$ and $\alpha$ for the deepest classifier are zero, which means the deepest classifier's supervision just comes from labels.

\section{Experiments\label{experiments}}
We evaluate self distillation on five convolutional neural networks (ResNet \cite{he2016deep}, WideResNet \cite{zagoruyko2016wide}, Pyramid ResNet~\cite{han2017deep}, ResNeXt \cite{xie2017aggregated}, VGG \cite{simonyan2014very}) and two datasets (CIFAR100 \cite{krizhevsky2009learning}, ImageNet \cite{deng2009imagenet}). Learning rate decay, $l_2$ regularizer and simple data argumentation are used during the training process.  All the experiments are implemented by PyTorch on GPU devices.

\subsection{Benchmark Datasets}
\textbf{CIFAR100:}
CIFAR100 dataset \cite{krizhevsky2009learning} consists of tiny (32x32 pixels) RGB images, has 100 classes and contains 50K images in training set and 10K images in testing set. Kernel sizes and strides of neural networks are adjusted to fit the size of tiny images.

\textbf{ImageNet:}
ImageNet2012 classification dataset \cite{deng2009imagenet} is composed of 1000 classes according to WordNet. Each class is depicted by thousands of images. We resize them into 256x256 pixels RGB images. Note that reported accuracy of ImageNet is computed on the validation set.

\begin{table*}[h]
\begin{center}
\begin{tabular}{|c|c|c|c|c|c|c|c|}
\hline
Neural Networks& Attribute &Baseline& Classifier 1/4 &Classifier 2/4 &Classifier3/4 &Classifier 4/4 & Ensemble\\
\cline{1-8}
\multirow{2}{*}{ResNet18}&Accuracy &77.09&67.23 & 73.80 &77.75 &78.38  &79.27 \\
\cline{2-8}
&Acceleration&1.00X & 3.11X &1.87X  &1.30X &1.00X  &0.93X \\
\cline{1-8}
\multirow{2}{*}{ResNet50}&Accuracy&77.68&67.87 &73.80&74.54&80.27&80.67\\
\cline{2-8}
&Acceleration&1.00X &4.64X&2.20X&1.23X&1.00X&0.93X\\
\cline{1-8}
\multirow{2}{*}{ResNet101}&Accuracy&77.98&68.17&75.43&80.98&81.01&81.72 \\
\cline{2-8}
&Acceleration&1.00X&9.00X &4.27X&1.11X&1.00X&0.96X\\
\cline{1-8}
\multirow{2}{*}{ResNet152}&Accuracy&79.21&68.84&78.22&81.43&81.61&82.29 \\
\cline{2-8}
&Acceleration&1.00X &13.36X&4.29X&1.07X&1.00X&0.98X\\
\cline{1-8}
\end{tabular}
\end{center}
\caption{Acceleration and accuracy (\%) for ResNet on CIFAR100.}
\label{table4}
\end{table*}

\subsection{Compared with Standard Training}
Results of experiments on CIFAR100 and ImageNet are displayed in Table \ref{table1} and Table \ref{imagenet} respectively. An ensemble result is obtained by simply adding the weighted outputs of the softmax layer in each classifier. It is observed that (i) all the neural networks benefit significantly from self distillation, with an increment of 2.65\% in CIFAR100 and 2.02\% in ImageNet on average. (ii) The deeper the neural networks are, the more improvement on performance they acquire, for example, an increment of 4.05\% in ResNet101 and 2.58\% in ResNet18. (iii) Generally speaking, naive ensemble works effectively on CIFAR100 yet with less and sometimes negative influence on ImageNet, which may be caused by the larger accuracy drop in shallow classifiers, compared with that on CIFAR100. (iv) Classifiers' depth plays a more crucial part in ImageNet, indicating there is less redundancy in neural networks for a complex task.

\subsection{Compared with Distillation}
Table \ref{table2} compares results of self distillation with that of five traditional distillation methods on CIFAR100 dataset. Here we focus on the accuracy boost of each method when the student models have the same computation and storage amount. From Table \ref{table2}, we make the following observations: (i) All the performance of distillation methods outperforms the directly trained student networks. (ii) Although self distillation doesn't have an extra teacher, it still outperforms most of the rest distillation methods.

One significant advantage of self distillation framework is that it doesn't need an extra teacher. In contrast, traditional distillation needs to design and train an over-parameterized teacher model at first. Designing a high quality teacher model needs tremendous experiments to find the best depth and architecture. In addition, training an over-parameterized teacher model takes much longer time. These problems can be directly avoided in self distillation, where both teachers and students models are sub-sections of itself. As depicted in Figure \ref{Compare}, 4.6X acceleration in training time can be achieved by self distillation compared with other distillation methods.

\subsection{Compared with Deeply Supervised Net}\label{compared with DSN}
The main difference between deeply supervised net and self distillation is that self distillation trains shallow classifiers from the deepest classifier's distillation instead of labels. The advantages can be seen in experiments, as shown in Table \ref{table3}, which compares the accuracy of each classifier in ResNet trained by deep supervision or self distillation on CIFAR100. The observations can be summarized as follows: (i) Self distillation outperforms deep supervision in every classifier. (ii) Shallow classifiers benefit more from self distillation.

The reasons for the phenomena are easy to understand.
In self distillation, (i) extra bottleneck layers are added to detect classifier-specific features, avoiding conflicts between shallow and deep classifiers. (ii) Distillation method has been employed in training the shallow classifiers instead of labels to boost the performance. (iii) Better shallow classifiers can obtain more discriminating features, which enhances the deeper classifiers performance in return.

\subsection{Scalable Depth for Adapting Inference}

Recently, a popular solution to accelerate convolutional neural networks is to design a scalable network, which means the depth or width of neural networks can change dynamically according to application requirements. For example, in the scenarios where response time is more important than accuracy, some layers or channels could be abandoned at runtime for acceleration \cite{yu2018slimmable}.

With a sharing backbone network, adaptive accuracy-acceleration tradeoff in inference becomes possible on resource-limited edge devices, which means that different depth classifiers can be automatically employed in applications according to dynamic accuracy demands in real word.
As can be observed in Table \ref{table4} that (i) three in four neural networks outperform their baselines by classifier 3/4, with an acceleration ratio of 1.2X on average. 3.16X acceleration ratio can be achieved with an accuracy loss at 3.3\% with classifier 2/4. (ii) Ensemble of the deepest three classifiers can bring 0.67\% accuracy improvement on average level with only 0.05\% penalty for computation, owing to that different classifiers share one backbone network.

\begin{figure}[htbp]
  \centering
  \includegraphics[width=8.38cm,height=4.13cm]{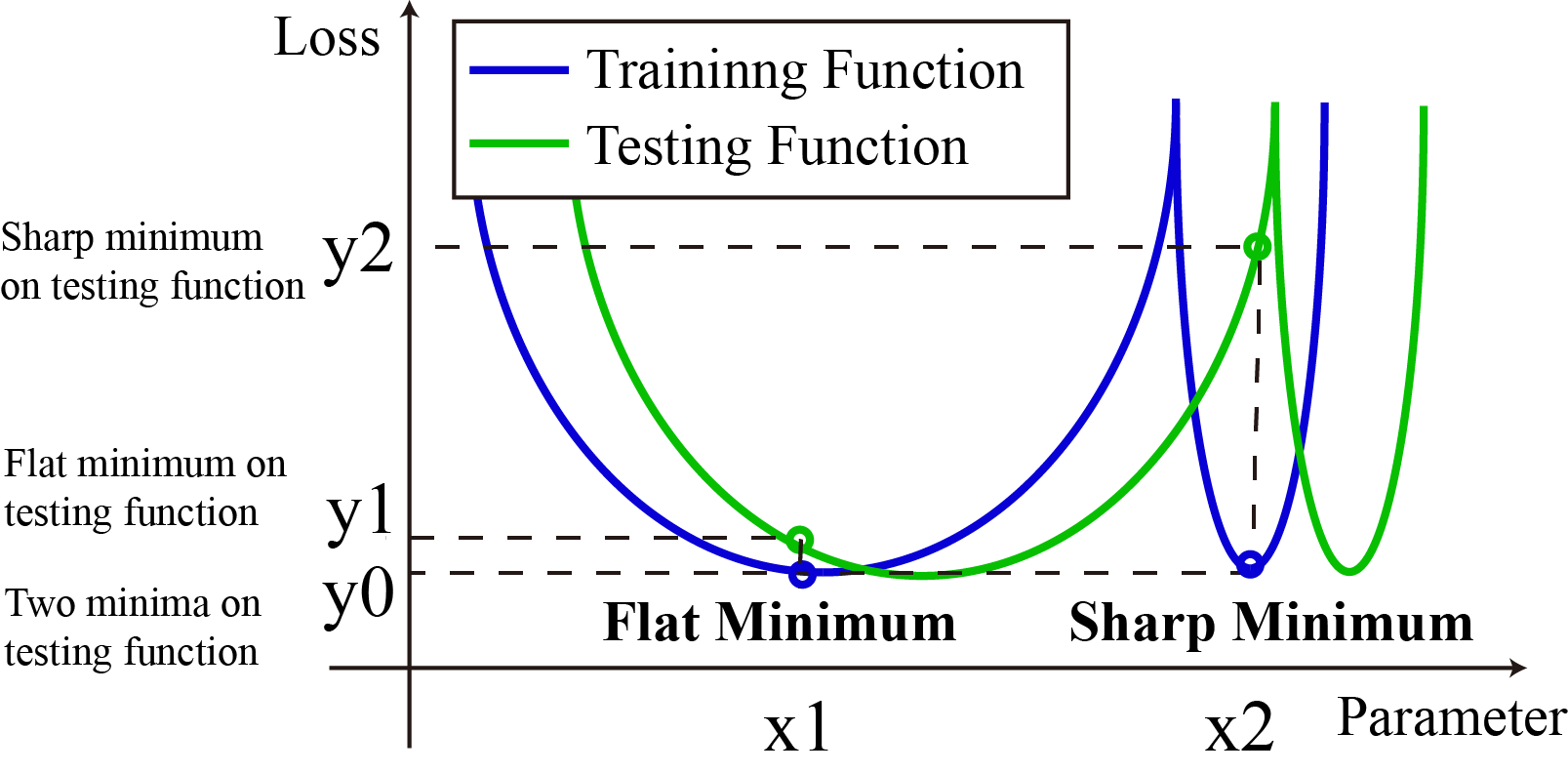}
  \caption{An intuitive explanation of the difference between flat and sharp minima \cite{keskar2016large}.}
  \label{FlatMinima}
\end{figure}

\section{Discussion and Future Works\label{discussion}}
In this section, we discuss the possible explanations of notable performance improvement brought by self distillation from perspectives of flat minima, vanishing gradients, and discriminating features, which will be followed by the section of future works for further improvement.

\begin{figure}[htbp]
  \centering
  \includegraphics[width=8.6cm,height=12cm]{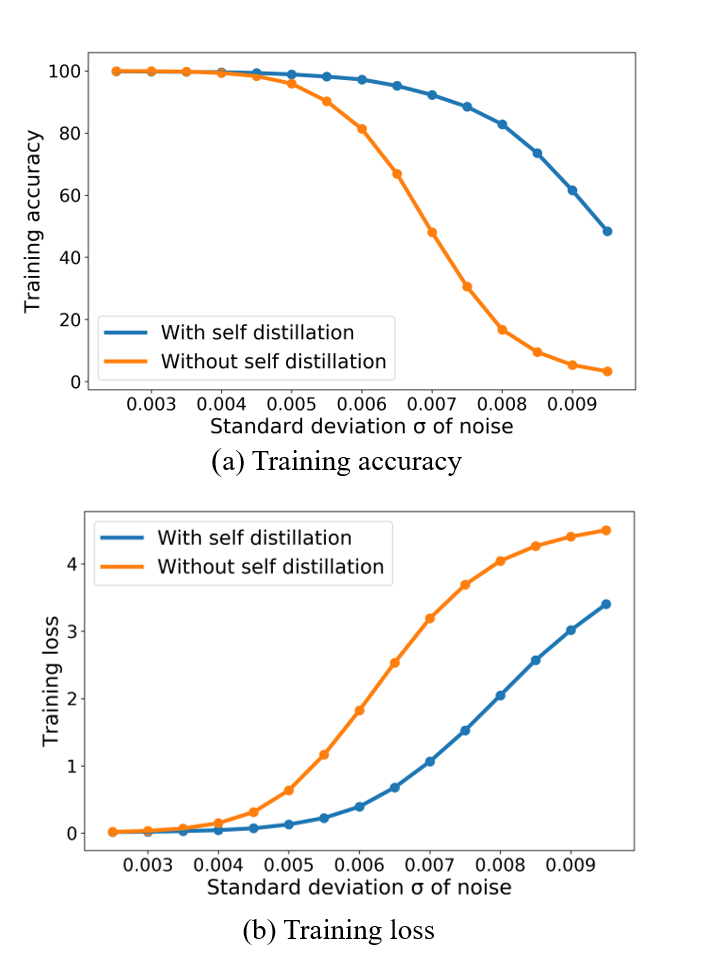}
  \caption{Comparison of training accuracy and loss with increasing Gaussian noise: models trained with self distillation are more tolerant to noise - flat minima.}
  \label{AddingNoise}
\end{figure}

\textbf{Self distillation can help models converge to flat minima which features in generalization inherently.}
It is universally acknowledged that although shallow neural networks (e.g. AlexNet) can also achieve almost zero loss on the training set, their performance on test set or in practical applications is far behind over-parameterized neural networks (e.g. ResNet) \cite{keskar2016large}.
Keskar \emph{et al.} proposed explanations that over-parameters models may converge easier to the flat minima, while shallow neural networks are more likely to be caught in the sharp minima, which is sensitive to the bias of data \cite{keskar2016large}.
Figure \ref{FlatMinima} gives an intuitive explanation of the difference between flat and sharp minima.
The X axis represents the parameters of models in one dimension.
The Y axis is the value of loss function.
The two curves denote the loss curves on training set and test set.
Both two minima (x1 for flat mimima and x2 for sharp minima) can achieve extremely small loss on the training set (y0).
Unfortunately, the training set and the test set are not independently and identically distributed.
While in the test, x1 and x2 are still utilized to find the minima y1 and y2 in the testing curve, which causes
severe bias in the sharp mimina curve (y2 - y0 is much larger than y1 - y0).

Inspired by the work of Zhang \emph{et al.} \cite{zhang2018deep}, we conduct the following experiments to show that the proposed self distillation framework can converge to a flat minimun.
Two 18-layer ResNets have been trained on CIFAR100 dataset firstly, one with self distillation and the other one not. Then Gaussian noise are added to the parameters of the two models and then their entropy loss and predicted accuracy on the training set are obtained and plotted in Figure \ref{AddingNoise}.
As can be seen in Figure \ref{AddingNoise}(a), the training set accuracy in the model trained with self distillation maintains at a very high level with noise level, presented as standard deviation of the Gaussian noises, keeping increasing.
While the training accuracy in the model without self distillation drops severely, as shown in Figure \ref{AddingNoise}(a).
Same observations and conclusions can be obtained in Figure \ref{AddingNoise}(b) with training loss as the metric.
Based on the above observations, we conclude that the models trained with self distillation are more flat.
According to the conclusion sourced from Figure \ref{FlatMinima}, the model trained with self distillation are more robust to perturbation of parameters.
Note that the 4/4 classifier is used in self distillation ResNet for a fair comparison.
To sum up, the model trained without self distillation is much more sensitive to the Gaussian noise. These experiments results support our view that self distillation helps models find flat minima, permitting better generalization performance.


\textbf{Self distillation prevents models from vanishing gradient problem.}
Due to vanishing gradient problem, very deep neural networks are hard to train, although they show better generalization performance.
In self distillation, the supervision on the neural networks is injected into different depth. It inherits the ability of DSN \cite{lee2015deeply} to address the vanishing gradient problem to some extent. Since the work of Lee \emph{et al.} \cite{lee2015deeply} has given the justification mathematically, we conduct the following experiments to support it.

\begin{figure}[htbp]
  \centering
  \includegraphics[width=8.73cm,height=3.31cm]{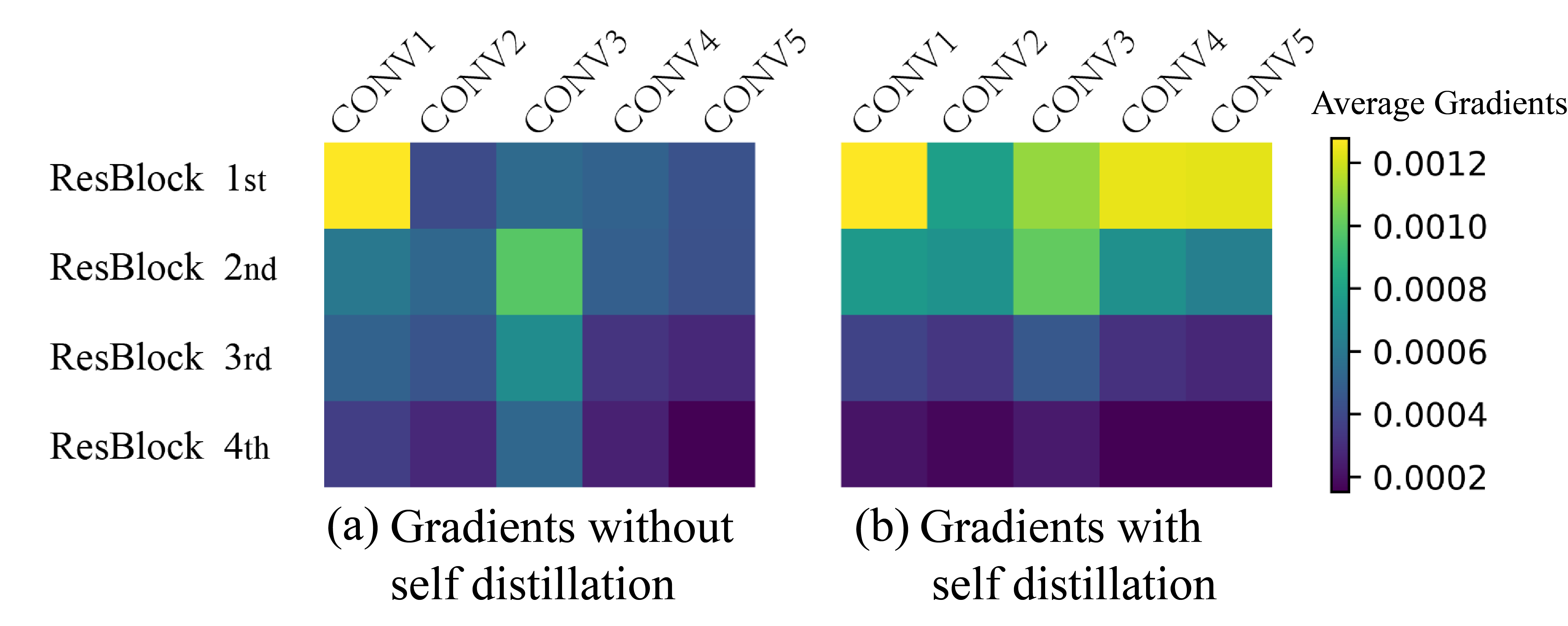}
  \caption{Statistics of layer-wised gradients.}
  \label{GradVanish}
\end{figure}

Two 18-layer ResNets are trained, one of them equipped with self distillation and the other one not. We compute the mean magnitude of gradients in each convolutional layer as shown in Figure \ref{GradVanish}.
It is observed that the magnitude of gradients of the model with self distillation (Figure \ref{GradVanish}(a)) is larger than the one without self distillation (Figure \ref{GradVanish}(b)), especially in the first and second ResBlocks.

\begin{figure}[h]
  \centering
  \includegraphics[width=8.33cm,height=7.134cm]{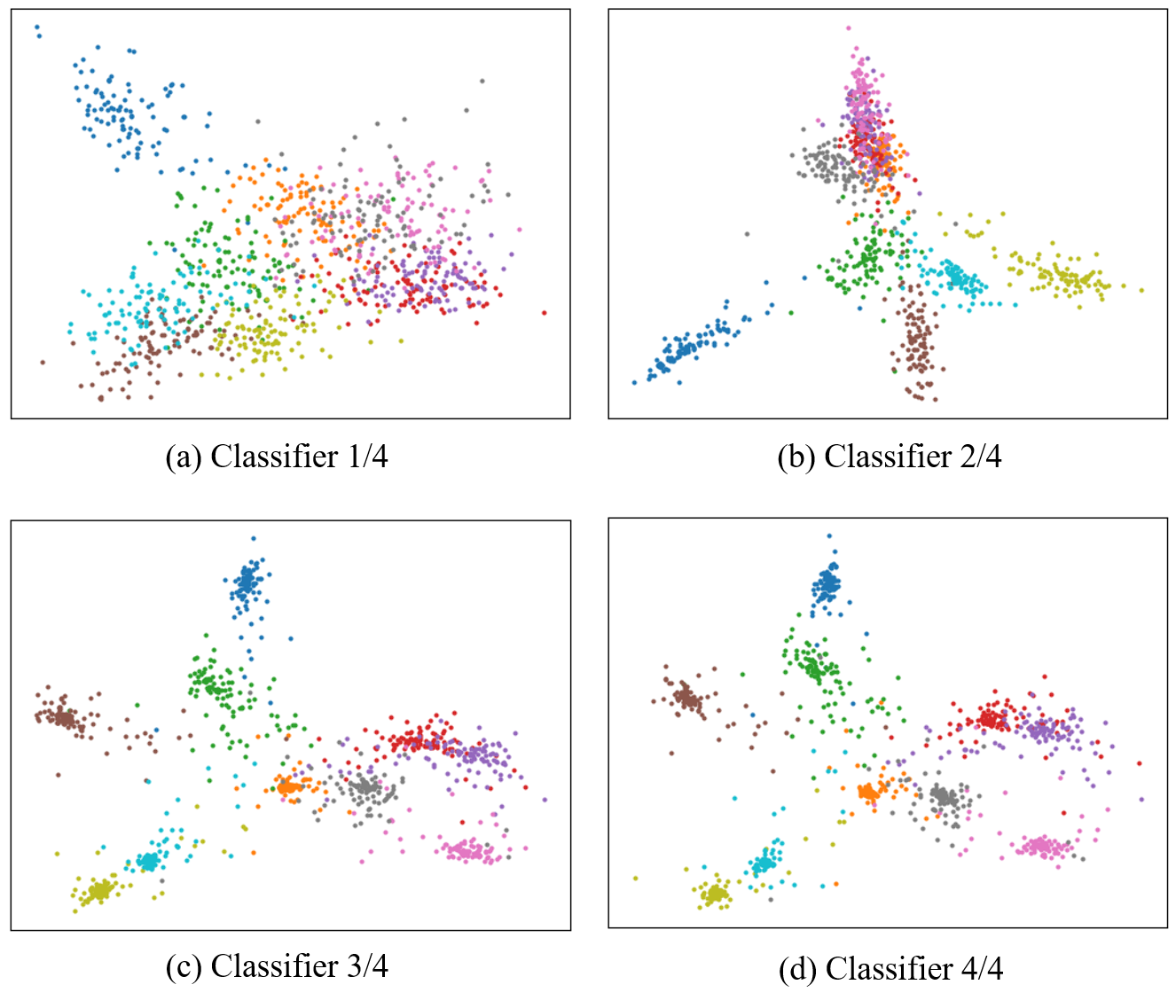}
  \caption{PCA (principal component analysis) visualization of feature distribution in four classifiers.}
  \label{FeatureDistri}
\end{figure}

\textbf{More discriminating features are extracted with deeper classifiers in self distillation.} Since there are multiple classifiers existing in self distillation, features of each classifier can be computed and analyzed to demonstrate their discriminating principle. As depicted in Figure \ref{FeatureDistri}, experiments on WideResNet trained on CIFAR100 are conducted to compare features of different classifiers.

Figure \ref{FeatureDistri} visualizes the distances of features in different classifiers.
To begin with, it is obvious that the deeper the classifier, the more concentrated clusters are observed.
In addition, the changes of the distances in shallow classifiers, as shown in Figure \ref{FeatureDistri}(a,b), are more severe than that in deep classifiers, as demonstrated in Figure \ref{FeatureDistri}(c,d).

\begin{table}[h]
\begin{center}
\begin{tabular}{|c|c|c|c|c|}
\hline
Classifier&SSE*&SSB**&SSE/SSB& Accuracy \\  \hline
Classifier1/4& 20.85 &1.08&19.21&71.21\\ \hline
Classifier2/4& 8.69&1.15&7.54 &80.86 \\ \hline
Classifier3/4& 11.42 &1.87&6.08&81.58 \\ \hline
Classifier4/4& 11.74 &2.05&5.73&81.59 \\ \hline
\end{tabular}
\end{center}
*SSE: Sum of squares due to error.

**SSB: Sum of squares between groups.
\caption{Measurement of sort separability and accuracy (\%) for each classifier on WideResNet28-12.}
\label{FeatureDistriDistance}
\end{table}

Table \ref{FeatureDistriDistance} further summarizes the sort separability for each classifier.
SSE stands for sum of squares due to error, and SSB is short for sum of squares between groups.
The smaller the SSE is, the denser the clusters are.
Also, the clusters become more discriminating with the SSB growing.
Here we use SSE/SSB to evaluate the distinct capability of the models. The smaller it is, the more clear the classifier is.
It can be seen in Table \ref{FeatureDistriDistance} that the SSE/SSB decreases as classifier goes deeper.
In summary, the more discriminating feature maps in the classifier, the higher accuracy the model achieves.

\textbf{Future Works}

\emph{Automatic adjustment of newly introduced hyper-parameters.}
To balance the loss of cross entropy, KL divergence, and hint loss, two hyper-parameters
$\lambda$ and $\alpha$ are introduced as shown in Equation \ref{equation5}.
Through the experiments, we find out that these two hyper-parameters have impacts on the performance. Due to limited computation resources, we have not done a through investigation.
In the near future, automatic adjustment of the two hyper-parameters can be explored using
learning rate decay like or momenta inspired algorithms.

\emph{Is the flat minimum found by self distillation ideal?}
Another unexplored domain is that we find a phenomenon during training that after the convergence of self distillation, continuing training of the deepest classifiers using conventional training method can further boost the performance from 0.3\% to 0.7\%, which are not included in all the Tables in the paper.
Despite that shallow classifiers help find the flat minimum, at the final stage of the training, they also prevent the deepest classifier from convergence.
Alternately switching between multiple training methods might further help the convergence.


\section{Conclusion\label{conclusion}}
We have proposed a novel training technique called self distillation and shown its advantage by comparing it with deeply supervised net and the previous distillation methods. This technique abandons the extra teacher model required in previous distillation methods and provides an adaptive depth architecture for time-accuracy tradeoffs at runtime. We also have explored the principle behind self distillation from the perspective of flat minima, gradients and discriminating feature.

Self distillation is more of a training technique to boost model performance rather than a method to compress nor accelerate models.
Although most of the previous research focuses on knowledge transfer among different models, we believe that knowledge transfer approaches inside one model like self distillation are also very promising.

{\small
\bibliographystyle{ieee}
\bibliography{egbib}

\begin{thebibliography}{10}\itemsep=-1pt

\bibitem{amthor2016impatient}
M.~Amthor, E.~Rodner, and J.~Denzler.
\newblock Impatient dnns-deep neural networks with dynamic time budgets.
\newblock In {\em British Machine Vision Conference}, 2016.

\bibitem{bagherinezhad2018label}
H.~Bagherinezhad, M.~Horton, M.~Rastegari, and A.~Farhadi.
\newblock Label refinery: Improving imagenet classification through label
  progression.
\newblock In {\em European Conference on Computer Vision}, 2018.

\bibitem{bahdanau2014neural}
D.~Bahdanau, K.~Cho, and Y.~Bengio.
\newblock Neural machine translation by jointly learning to align and
  translate.
\newblock In {\em International Conference on Medical Image Computing and
  Computer-Assisted Intervention}, 2015.

\bibitem{buciluǎ2006model}
C.~Buciluǎ, R.~Caruana, and A.~Niculescu-Mizil.
\newblock Model compression.
\newblock In {\em Proceedings of the 12th ACM SIGKDD international conference
  on Knowledge discovery and data mining}, pages 535--541. ACM, 2006.

\bibitem{courbariaux2015binaryconnect}
M.~Courbariaux, Y.~Bengio, and J.-P. David.
\newblock Binaryconnect: Training deep neural networks with binary weights
  during propagations.
\newblock In {\em Advances in neural information processing systems}, pages
  3123--3131, 2015.

\bibitem{deng2009imagenet}
J.~Deng, W.~Dong, R.~Socher, L.-J. Li, K.~Li, and L.~Fei-Fei.
\newblock Imagenet: A large-scale hierarchical image database.
\newblock In {\em Computer Vision and Pattern Recognition}, pages 248--255.
  Ieee, 2009.

\bibitem{dou20163d}
Q.~Dou, H.~Chen, Y.~Jin, L.~Yu, J.~Qin, and P.-A. Heng.
\newblock 3d deeply supervised network for automatic liver segmentation from ct
  volumes.
\newblock In {\em International Conference on Medical Image Computing and
  Computer-Assisted Intervention}, pages 149--157. Springer, 2016.

\bibitem{fu2017look}
J.~Fu, H.~Zheng, and T.~Mei.
\newblock Look closer to see better: Recurrent attention convolutional neural
  network for fine-grained image recognition.
\newblock In {\em Proceedings of the IEEE Conference on Computer Vision and
  Pattern Recognition}, volume~2, page~3, 2017.

\bibitem{furlanello2018born}
T.~Furlanello, Z.~C. Lipton, M.~Tschannen, L.~Itti, and A.~Anandkumar.
\newblock Born again neural networks.
\newblock In {\em International Machine Learning Conference}, 2018.

\bibitem{gupta2016cross}
S.~Gupta, J.~Hoffman, and J.~Malik.
\newblock Cross modal distillation for supervision transfer.
\newblock In {\em Computer Vision and Pattern Recognition}, pages 2827--2836,
  2016.

\bibitem{han2017deep}
D.~Han, J.~Kim, and J.~Kim.
\newblock Deep pyramidal residual networks.
\newblock In {\em Computer Vision and Pattern Recognition (CVPR), 2017 IEEE
  Conference on}, pages 6307--6315. IEEE, 2017.

\bibitem{han2015deep}
S.~Han, H.~Mao, and W.~J. Dally.
\newblock Deep compression: Compressing deep neural networks with pruning,
  trained quantization and huffman coding.
\newblock 2016.

\bibitem{han2015learning}
S.~Han, J.~Pool, J.~Tran, and W.~Dally.
\newblock Learning both weights and connections for efficient neural network.
\newblock In {\em Advances in neural information processing systems}, pages
  1135--1143, 2015.

\bibitem{he2016deep}
K.~He, X.~Zhang, S.~Ren, and J.~Sun.
\newblock Deep residual learning for image recognition.
\newblock In {\em Proceedings of the IEEE conference on computer vision and
  pattern recognition}, pages 770--778, 2016.

\bibitem{hinton2015distilling}
G.~Hinton, O.~Vinyals, and J.~Dean.
\newblock Distilling the knowledge in a neural network.
\newblock In {\em Advances in neural information processing systems}, 2014.

\bibitem{howard2017mobilenets}
A.~G. Howard, M.~Zhu, B.~Chen, D.~Kalenichenko, W.~Wang, T.~Weyand,
  M.~Andreetto, and H.~Adam.
\newblock Mobilenets: Efficient convolutional neural networks for mobile vision
  applications.
\newblock In {\em Computer Vision and Pattern Recognition (CVPR), 2017 IEEE
  Conference on}, 2017.

\bibitem{huang2017multi}
G.~Huang, D.~Chen, T.~Li, F.~Wu, L.~van~der Maaten, and K.~Q. Weinberger.
\newblock Multi-scale dense networks for resource efficient image
  classification.
\newblock In {\em International Conference on Medical Image Computing and
  Computer-Assisted Intervention}, 2017.

\bibitem{huang2016deep}
G.~Huang, Y.~Sun, Z.~Liu, D.~Sedra, and K.~Q. Weinberger.
\newblock Deep networks with stochastic depth.
\newblock In {\em European conference on computer vision}, pages 646--661.
  Springer, 2016.

\bibitem{iandola2016squeezenet}
F.~N. Iandola, S.~Han, M.~W. Moskewicz, K.~Ashraf, W.~J. Dally, and K.~Keutzer.
\newblock Squeezenet: Alexnet-level accuracy with 50x fewer parameters and< 0.5
  mb model size.
\newblock In {\em International Conference on Learning Representations}, 2016.

\bibitem{keskar2016large}
N.~S. Keskar, D.~Mudigere, J.~Nocedal, M.~Smelyanskiy, and P.~T.~P. Tang.
\newblock On large-batch training for deep learning: Generalization gap and
  sharp minima.
\newblock In {\em International Conference on Learning Representations}, 2017.

\bibitem{krizhevsky2009learning}
A.~Krizhevsky and G.~Hinton.
\newblock Learning multiple layers of features from tiny images.
\newblock Technical report, Citeseer, 2009.

\bibitem{krizhevsky2012imagenet}
A.~Krizhevsky, I.~Sutskever, and G.~E. Hinton.
\newblock Imagenet classification with deep convolutional neural networks.
\newblock In {\em Advances in neural information processing systems}, pages
  1097--1105, 2012.

\bibitem{kuen2018stochastic}
J.~Kuen, X.~Kong, Z.~Lin, G.~Wang, J.~Yin, S.~See, and Y.-P. Tan.
\newblock Stochastic downsampling for cost-adjustable inference and improved
  regularization in convolutional networks.
\newblock In {\em Proceedings of the IEEE Conference on Computer Vision and
  Pattern Recognition}, pages 7929--7938, 2018.

\bibitem{lee2015deeply}
C.-Y. Lee, S.~Xie, P.~Gallagher, Z.~Zhang, and Z.~Tu.
\newblock Deeply-supervised nets.
\newblock In {\em Artificial Intelligence and Statistics}, pages 562--570,
  2015.

\bibitem{lin2017feature}
T.-Y. Lin, P.~Doll{\'a}r, R.~Girshick, K.~He, B.~Hariharan, and S.~Belongie.
\newblock Feature pyramid networks for object detection.
\newblock In {\em Proceedings of the IEEE Conference on Computer Vision and
  Pattern Recognition}, volume~1, page~4, 2017.

\bibitem{lin2018focal}
T.-Y. Lin, P.~Goyal, R.~Girshick, K.~He, and P.~Doll{\'a}r.
\newblock Focal loss for dense object detection.
\newblock {\em IEEE transactions on pattern analysis and machine intelligence},
  2018.

\bibitem{liu2018ktan}
P.~Liu, W.~Liu, H.~Ma, T.~Mei, and M.~Seok.
\newblock Ktan: Knowledge transfer adversarial network.
\newblock In {\em Association for the Advance of Artificial Intelligence},
  2019.

\bibitem{liu2016ssd}
W.~Liu, D.~Anguelov, D.~Erhan, C.~Szegedy, S.~Reed, C.-Y. Fu, and A.~C. Berg.
\newblock Ssd: Single shot multibox detector.
\newblock In {\em European conference on computer vision}, pages 21--37.
  Springer, 2016.

\bibitem{mnih2014recurrent}
V.~Mnih, N.~Heess, A.~Graves, et~al.
\newblock Recurrent models of visual attention.
\newblock In {\em Advances in neural information processing systems}, pages
  2204--2212, 2014.

\bibitem{papernot2016distillation}
N.~Papernot, P.~McDaniel, X.~Wu, S.~Jha, and A.~Swami.
\newblock Distillation as a defense to adversarial perturbations against deep
  neural networks.
\newblock In {\em 2016 IEEE Symposium on Security and Privacy}, pages 582--597.
  IEEE, 2016.

\bibitem{rastegari2016xnor}
M.~Rastegari, V.~Ordonez, J.~Redmon, and A.~Farhadi.
\newblock Xnor-net: Imagenet classification using binary convolutional neural
  networks.
\newblock In {\em European Conference on Computer Vision}, pages 525--542.
  Springer, 2016.

\bibitem{romero2014fitnets}
A.~Romero, N.~Ballas, S.~E. Kahou, A.~Chassang, C.~Gatta, and Y.~Bengio.
\newblock Fitnets: Hints for thin deep nets.
\newblock In {\em International Conference on Learning Representations}, 2015.

\bibitem{shen2018meal}
Z.~Shen, Z.~He, and X.~Xue.
\newblock Meal: Multi-model ensemble via adversarial learning.
\newblock In {\em Association for the Advance of Artificial Intelligence},
  2019.

\bibitem{simonyan2014very}
K.~Simonyan and A.~Zisserman.
\newblock Very deep convolutional networks for large-scale image recognition.
\newblock 2015.

\bibitem{veit2017convolutional}
A.~Veit and S.~Belongie.
\newblock Convolutional networks with adaptive inference graphs.
\newblock In {\em European conference on computer vision}, 2018.

\bibitem{wang2018skipnet}
X.~Wang, F.~Yu, Z.-Y. Dou, T.~Darrell, and J.~E. Gonzalez.
\newblock Skipnet: Learning dynamic routing in convolutional networks.
\newblock In {\em Proceedings of the European Conference on Computer Vision
  (ECCV)}, pages 409--424, 2018.

\bibitem{wu2018blockdrop}
Z.~Wu, T.~Nagarajan, A.~Kumar, S.~Rennie, L.~S. Davis, K.~Grauman, and
  R.~Feris.
\newblock Blockdrop: Dynamic inference paths in residual networks.
\newblock In {\em Proceedings of the IEEE Conference on Computer Vision and
  Pattern Recognition}, pages 8817--8826, 2018.

\bibitem{xie2017aggregated}
S.~Xie, R.~Girshick, P.~Doll{\'a}r, Z.~Tu, and K.~He.
\newblock Aggregated residual transformations for deep neural networks.
\newblock In {\em Computer Vision and Pattern Recognition (CVPR), 2017 IEEE
  Conference on}, pages 5987--5995. IEEE, 2017.

\bibitem{yu2018slimmable}
J.~Yu, L.~Yang, N.~Xu, J.~Yang, and T.~Huang.
\newblock Slimmable neural networks.
\newblock In {\em International Conference on Learning Representations}, 2019.

\bibitem{yu2017volumetric}
L.~Yu, X.~Yang, H.~Chen, J.~Qin, and P.-A. Heng.
\newblock Volumetric convnets with mixed residual connections for automated
  prostate segmentation from 3d mr images.
\newblock In {\em Association for the Advance of Artificial Intelligence},
  pages 66--72, 2017.

\bibitem{zagoruyko2016wide}
S.~Zagoruyko and N.~Komodakis.
\newblock Wide residual networks.
\newblock In {\em British Machine Vision Conference}, 2016.

\bibitem{zagoruyko2016paying}
S.~Zagoruyko and N.~Komodakis.
\newblock Paying more attention to attention: Improving the performance of
  convolutional neural networks via attention transfer.
\newblock In {\em International Conference on Learning Representations}, 2017.

\bibitem{zhang2018deep}
Y.~Zhang, T.~Xiang, T.~M. Hospedales, and H.~Lu.
\newblock Deep mutual learning.
\newblock In {\em Proceedings of the IEEE Conference on Computer Vision and
  Pattern Recognition}, pages 4320--4328, 2018.

\end{thebibliography}
}

\end{document}